\pdfoutput=1
\PassOptionsToPackage{hyphens}{url}
\documentclass[11pt]{article}
\usepackage{naacl2021}

\usepackage{times}
\usepackage{makecell}
\usepackage{latexsym}
\usepackage{amsfonts}
\usepackage{url}
\usepackage[normalem]{ulem}
\usepackage{colortbl}

\usepackage[T1]{fontenc}
\usepackage[utf8]{inputenc}
\usepackage{graphicx}
\usepackage{microtype}

\makeatletter
\newcommand{\thickhline}{%
    \noalign {\ifnum 0=`}\fi \hrule height 1pt
    \futurelet \reserved@a \@xhline
}
\makeatother

\title{\textsc{MediaSum}: A Large-scale Media Interview Dataset for Dialogue Summarization}

\author{Chenguang Zhu\thanks{$\;\;$Equal contribution}$\;$, Yang Liu$^*$, Jie Mei, Michael Zeng \\
Microsoft Cognitive Services Research Group\\
  \texttt{\{chezhu,yaliu10,jimei,nzeng\}@microsoft.com} \\}

\begin{document}
\maketitle
\begin{abstract}
This paper introduces \textsc{MediaSum}\footnote{\url{https://github.com/zcgzcgzcg1/MediaSum/}}, a large-scale media interview dataset consisting of 463.6K transcripts with abstractive summaries. 
To create this dataset, we collect interview transcripts from NPR and CNN and employ the overview and topic descriptions as summaries. Compared with existing public corpora for dialogue summarization, our dataset is an order of magnitude larger and contains complex multi-party conversations from multiple domains.
We conduct statistical analysis to demonstrate the unique positional bias exhibited in the transcripts of televised and radioed interviews.
We also show that \textsc{MediaSum} can be used in transfer learning to improve a model's performance on other dialogue summarization tasks.

\end{abstract}

\section{Introduction}
Dialogue summarization can provide a succinct synopsis for conversations between two or more participants, based on human-transcribed or machine-generated transcripts. 
Dialogue summaries are useful for participants to recap salient information in the talk and for absentees to grasp the key points. 
As a result, several models have been recently proposed to summarize daily conversations \citep{samsum,chen2020multi}, meeting transcripts \citep{hmnet} and customer support conversations \citep{didi}.

However, compared with the abundance of text summarization datasets, there are very few public datasets for dialogue summarization. And existing datasets are limited to their small sizes. 
For example, the benchmark datasets for meeting summarization, AMI \citep{ami} and ICSI \citep{icsi}, only contain transcripts and abstractive summaries for 137 and 59 business meetings, respectively. 
While recently some larger dialogue summarization datasets have been proposed, they are either built from a narrow domain, e.g. the CRD3 dataset \cite{crd3} which is built from conversations in a live-streamed show for the Dungeons and Dragons game, or not publicized due to privacy reasons, e.g. the Didi dataset \citep{didi} from customer service conversations. This lack of large-scale dialogue summarization datasets is due to a higher labeling cost compared with news articles and privacy issues with many real daily dialogues and business meetings.

On the other hand, media interview transcripts and the associated summaries/topics can be a valuable source for dialogue summarization. In a broadcast interview, the host discusses various topics with one or more guests. As many interviews proceed with pre-defined topics, the accompanying summaries are of a relatively high quality. Also, the wide variety of topics, different backgrounds of speakers, and the colloquial form of chat make these interviews very close to daily conversations and business meetings.

Therefore, we collect public interview transcripts and the associated summaries/topics from NPR and CNN to build a large-scale dialogue summarization dataset, \textsc{MediaSum}.

In NPR, each transcript comes with an overview of the interview, which is used as the summary in our dataset. We leverage the INTERVIEW dataset \citep{interview} to get transcripts and crawl the associated descriptions. We end up with 49.4K NPR transcripts with summaries.

We then collect 269.4K CNN interview transcripts from 2000 to 2020, each with a list of topic descriptions. As many CNN interviews contain multiple topics, we conduct segmentation at the boundary of commercial breaks to assign each topic to the most relevant interview segment via lexical matching. %Specifically, we assign each topic to the segment with most shared non-stop words. 
In this way, we not only obtain transcripts with a more concentrated topic but also enlarge the total number of instances. We end up with 414.2K CNN transcript segments with topic descriptions as summaries. Thus, in total, our \textsc{MediaSum} dataset contains 463.6K transcripts with summaries.

We show that compared to existing public dialogue summarization datasets, \textsc{MediaSum} contains more speakers, longer conversation and is an order of magnitude larger. Also, we demonstrate the unique positional bias in interview dialogues: while a televised interview often mentions keywords in the summary at the beginning of the program, a radio interview usually mentions these keywords at both the beginning and the end of the program. %This novel finding infers that discourse analysis should take transcript source into consideration. Also, it shows that our dataset covers a variety of transcripts with different positional bias.

In experiments, we evaluate several benchmark summarization models on our dataset. We then show that after fine-tuning on \textsc{MediaSum}, models' performance can be improved on other dialogue summarization tasks like AMI, ICSI and SAMSum, demonstrating the transfer learning capability of our dataset.

\section{Related Work}
Due to the success of corpus-based methods, the past decade saw the emergence of many dialogue datasets on various domains \cite{multiwoz,ubuntu}. However, very few of these datasets contain corresponding summary text. As human dialogues have very different structures and language patterns from written articles, dialogue summarization models can only limitedly benefit from the largely available news summarization data \citep{hmnet}. 

Current public datasets for dialogue summarization are either very small or in a specific domain. AMI \citep{ami} and ICSI \citep{icsi} contain 137 and 59 meeting transcripts with abstractive summaries. %, from a total of 100 and 70 hours of recording. 
AMI meetings are recorded in an artificial environment with actors and ICSI contains meetings of a speech group. MultiWOZ \citep{multiwoz} is a multi-domain task-oriented dialogue dataset where the instructions have been used as summaries \citep{summonmultiwoz}. All dialogues are conducted between one user and one agent on the topic of booking and inquiry. SAMSum \citep{samsum} hires linguists to write messenger-like daily conversations. Although the dialogues are open-domain, they are not from real human conversations. CRD3 \citep{crd3} contains 159 episodes from the Critical Role show with transcribed conversations between Dungeons and Dragon players. %ScriptBase \citep{movie} includes 1,276 movie scripts with summaries. However, it contains a lot of plot descriptions and script notes instead of spoken dialogues. 
Additionally, there are non-public dialogue summarization datasets in th domains of customer support \citep{didi} and medical conversation \citep{krishna2020generating}.

%Compared to these corpora, our \textsc{MediaSum} dataset is from real multi-party interviews, covers a wide variety of domains and is an order of magnitude larger.

\section{Media Interview Dataset: \textsc{MediaSum}}
\subsection{Data collection}
We first collect interview transcriptions from National Public Radio (NPR, \url{www.npr.org}). The INTERVIEW dataset \citep{interview} contains 105K transcripts from NPR but does not include interview summaries or the link to the transcript page. We find a majority of NPR interviews come with an overview description before the transcription text, which can be used as summaries. Thus, for each interview in the INTERVIEW dataset, we use the NPR searching service to get the link to the corresponding page and extract the description text if it exists. We filter out descriptions with more than 200 words and collect 49.4K transcripts with summaries.

The CNN transcription service provides transcripts of televised interviews and a list of discussed topics, which can be used as summaries (\url{transcripts.cnn.com}). We crawl CNN transcripts from 2014 to 2020, combined with the data from 2000 to 2014 \citep{cnn2000_2014}, and end up with 269.4K transcripts with summaries.

\textbf{Transcript segmentation for topic match.} Interviews with multiple topics are often long, and the mixing of multiple topics makes it hard for models to generate accurate summaries. Among the collected CNN interviews, 157.9K transcripts, or 58.6\%, have more than one topic. Thus, we try to partition multi-topic interviews into segments and match each topic to a segment. We find that the televised CNN interviews often contain several commercial breaks marked in the transcript. These ads usually come in between topics. Therefore, we partition the transcript at the boundaries of commercial breaks. Then, we assign each topic to the segment containing the most (at least one) non-stop words in the topic. We do not count the last 50 words in a segment where the host often reminds watchers of the next topic after the commercial break. %In this way, many segments with a concentrated theme are associated with a unique topic. 
Among the 157.9K multi-topic interviews, 330.4K segments are associated with at least one topic.
%, 177.4K out of which has a single associated topic.
To make sure that the summary contains enough information, we filter out summaries with fewer than 5 words. In the end, we construct 414.2K CNN interview transcripts with summaries.

As transcripts from the NPR and CNN are from similar domains, we combine them into a unified summarization dataset, \textsc{MediaSum}, containing 463.6K pairs of transcripts and summaries. As far as we know, this is the largest public open-domain dialogue summarization dataset. We show an example dialogue with its summary in Table~\ref{tab:example}.

Here, we note that the summary styles of NPR and CNN are different. Table~\ref{tab:cnnnpr_stat} shows that although the dialogue length and number of speakers are similar in NPR and CNN, the summaries from NPR are much longer and more abstractive, indicated by a higher ratio of novel words in summary that do not appear in the dialogue.

\begin{table}[t]
    \centering
    \begin{tabular}{l|c|c}
    \thickhline
        \textbf{Statistics} & NPR & CNN
        \\
        \hline
        Dialogues & 49,420 & 414,176\\
        Avg. words in dialogue & 906.3 & 1,630.9\\
        Avg. words in summary & 40.2 & 11.3\\
        Turns & 24.2 & 30.7\\
        Speakers & 4.0 & 6.8\\
        Novel summary words & 33.6\% & 24.9\%\\
        \thickhline
    \end{tabular}
    \caption{Data statistics of NPR and CNN transcripts and summaries.\label{tab:cnnnpr_stat}}
\end{table}

\subsection{Data statistics}
In this section, we investigate different aspects of the \textsc{MediaSum} dataset via statistics. 

We leverage the Latent Dirichlet Allocation \citep{lda} tool in scikit-learn package \citep{scikit-learn} to analyze the main dialogue topics. We manually name the topic clusters based on the returned top 10 words in each cluster. The top 5 topics are politics (26.3\%), international news (13.3\%), crime (12.7\%), economy (12.5\%) and US news (11.7\%).

The dialogues in \textsc{MediaSum} have on average 30.0 turns, 6.5 speakers and 1,553.7 words, and the summaries have on average 14.4 words. This shows that most dialogues in our dataset are multi-party conversations of medium to long lengths. %A distribution plot is shown in Appendix~\ref{sec:appendix_stat}.
% The multi-turn structure and large quantity of tokens also make the task harder than traditional news summarization tasks. 

Table~\ref{tab:stat} compares \textsc{MediaSum} with other public dialogue summarization datasets. As shown, \textsc{MediaSum} contains much longer dialogues and more speakers than MultiWOZ 2.0 and SAMSum. This makes it suitable for training models targeted for multi-party dialogue or meeting summarization. Also, while AMI, ICSI and MultiWOZ 2.0 contain dialogues either from limited domains or under artificial context, \textsc{MediaSum} is a much larger dataset containing radioed and televised interview transcripts covering much broader topics.

\begin{table*}[h]
    \centering
    \begin{tabular}{l|c|c|c|c|c|c|c}
    \thickhline
        Dataset & \textsc{MediaSum} & AMI & ICSI & DiDi & CRD3 & MultiWOZ & SAMSum
        \\
        \hline
        Source & \multicolumn{5}{c|}{Transcribed Speech} & \multicolumn{2}{c}{Written}\\
        \hline
        Type & Interview & Meeting & Meeting & Customer & Game & Booking & Daily\\
        Real dialogue & $\checkmark$ & $\checkmark$ & $\checkmark$ & $\checkmark$ & $\checkmark$ & $\checkmark$ & $\times$ \\
        Open domain & $\checkmark$ & $\times$ & $\times$ & $\times$ & $\times$ & $\times$ & $\checkmark$ \\
        Public & $\checkmark$ & $\checkmark$ & $\checkmark$ & $\times$ & $\checkmark$ & $\checkmark$ & $\checkmark$ \\
        \hline
        Dialogues & 463,596 & 137 & 59 & 328,880 & 159 & 10,438 & 16,369\\
        Dial. words & 1,553.7 & 4,757 & 10,189 & / & 31,802.8 & 180.7 & 83.9\\
        Summ. words & 14.4 & 322 & 534 & / & 2062.3 & 91.9 & 20.3\\
        Turns & 30.0 & 289 & 464 & / & 2,507.4 & 13.7 & 9.9\\
        Speakers & 6.5 & 4 & 6.2 & 2 & 9.6 & 2 & 2.2\\
        \thickhline
    \end{tabular}
    \caption{Comparison of dialogue summarization datasets. The number of dialogue words, summary words, turns and speakers are all averaged across all dialogues in the dataset.\label{tab:stat}}
\end{table*}

\subsection{Positional Bias}
\label{sec:pos}
It has been found that in many news articles, the most important information is often shown at the beginning, i.e. the inverted pyramid structure \citep{top3}. In this section, we investigate whether a similar positional bias is present in multi-party dialogues.

\begin{figure}[t]
    \centering
    %\vspace{-0.5cm}
    \includegraphics[trim=0cm 0cm 0cm 0cm, clip, width=0.95\linewidth]{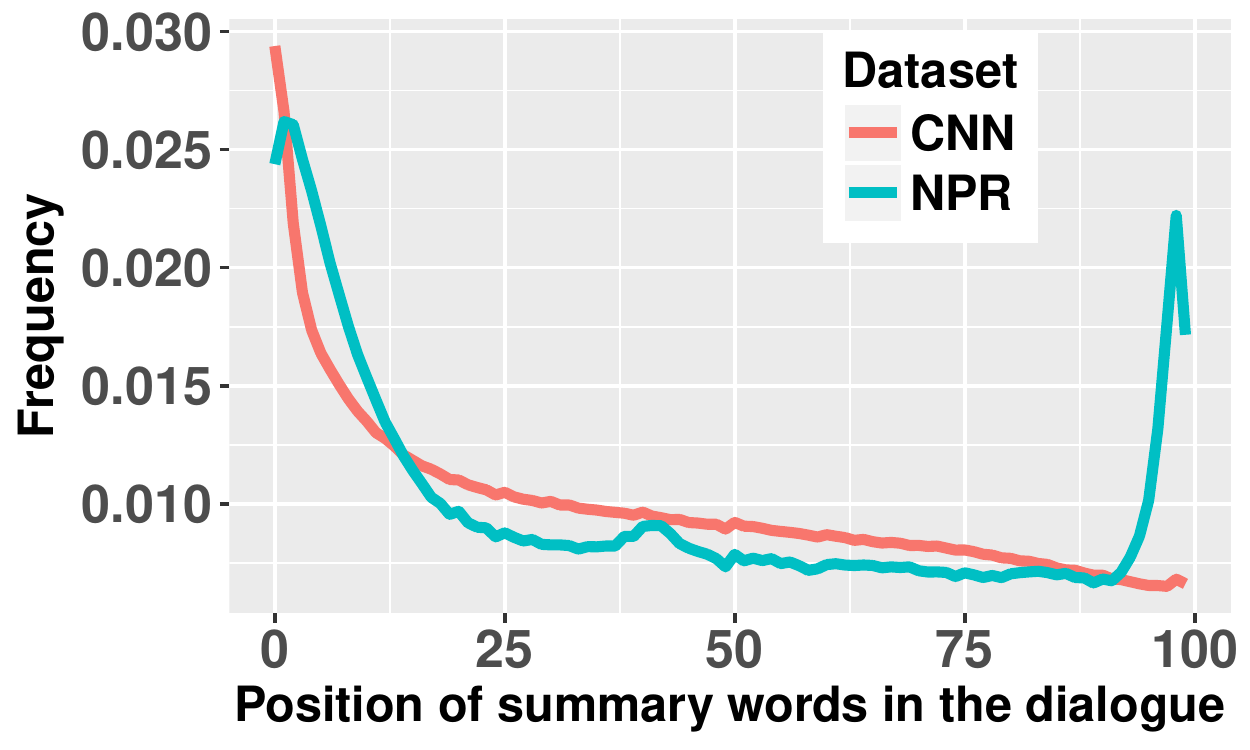}
    %\vspace{-4cm}
    \caption{The frequency of the non-stop summary words appearing at different positions of the dialogue. The positions are normalized to [0, 100].}
    \label{fig:pos}
\end{figure}

We record the position of each non-stop word in the transcript that also appears in the summary.
To normalize, we partition each transcript into 100 equal-length bins and count the frequency that summary words appear in each bin. As shown in Fig.~\ref{fig:pos}, similar to news articles, the beginning of transcripts from both CNN and NPR contain more summary words on average. %This corresponds to the anchor introducing the main topic to the listeners and viewers.
However, different from televised CNN interviews, NPR programs also contain many summary words near the end. To make sure that the trend in CNN is not caused by topic segmentation, we compute the frequency for original single-topic CNN transcripts and find that the trend is very similar to the overall distribution (Appendix~\ref{sec:appendix_posbias}). Thus, we suggest that the difference in positional bias between televised and radioed programs may be because viewers watching interviews on TV are relatively more focused, diminishing the need to recapitulate the main points before the program ends.

%Moreover, as CNN and NPR transcripts exhibit different positional bias, the \textsc{MediaSum} is suitable for training models for different dialogue domains with various positional bias.

\section{Experiments}
\subsection{Results on MediaSum}
We apply several benchmark summarization models to the \textsc{MediaSum} dataset and report the results, including PTGen \citep{pgnet}, the pre-trained models UniLM-base-uncased \citep{UniLM} and BART-Large \citep{bart}. The input concatenates transcripts from all turns, each prepended with the speaker name. %We truncate the input after the first 1,024 tokens. 
We also include the LEAD-3 baseline which takes the first three sentences of the transcript as the summary. More implementation details are shown in Appendix~\ref{sec:appendix_imp}.

We randomly select 10K instances for validation and another 10K for test. We use the ROUGE \citep{rouge} metrics and hyper-parameters are chosen based on the highest ROUGE-L score on the validation set.

As shown in Table~\ref{tab:res},  
the LEAD-3 baseline has a relatively weak performance, indicating that media dialogues exhibit less lead bias than news articles. This aligns with the general guideline to avoid inverted pyramid structure in digital programs \citep{abandoninvpyr}.
Moreover, pre-trained models such as BART and UniLM outperform the non-pre-trained PTGen model, showing the effectiveness of pre-training.%a similar trend as in news summarization \citep{pgnet}.

\begin{table}[t]
    \centering
    \begin{tabular}{l|c|c|c}
    \thickhline
        \textbf{Model} & \textbf{R-1} & \textbf{R-2} & \textbf{R-L}
        \\
        \thickhline
        LEAD-3 & 14.96 & 5.10 & 13.29 \\
        %LEAD-2-1 & 13.25 & 4.35 & 11.80 \\
        \hline
        PTGen & 28.77 & 12.24 & 24.18 \\
       % BART & 35.13 & 18.00 & 31.97\\
       UniLM &32.70&17.27&29.82\\
              BART & \textbf{35.09}&\textbf{18.05}&\textbf{31.44}\\
        \thickhline
    \end{tabular}
    \caption{ROUGE-1, ROUGE-2 and ROUGE-L F1 scores for models on \textsc{MediaSum} test set.\label{tab:res}}
\end{table}

\subsection{Transfer Learning}
In this section, we evaluate the transfer capability of \textsc{MediaSum} by employing it for further training to improve the performance on other dialogue summarization tasks of different domains and styles.
Specifically, we take the pre-trained model UniLM \citep{UniLM}, fine-tune it on \textsc{MediaSum}, and then train it on datasets for meeting and dialogue summarization: AMI \cite{ami}, ICSI \cite{icsi} and SAMSum \cite{samsum}.

As shown in Table~\ref{tab:transfer}, on all three datasets, training on \textsc{MediaSum} leads to improvement on the target dataset. This shows the potential of using \textsc{MediaSum} as a transfer learning dataset for other dialogue summarization tasks.

\begin{table}[]
\begin{tabular}{l|c|c|c}
\thickhline
\textbf{Model}             & \textbf{R-1}    & \textbf{R-2}    & \textbf{R-L}    \\ \thickhline
\rowcolor[gray]{0.95}
\multicolumn{4}{c}{AMI}                                        \\ \thickhline
UniLM             & 50.61 & \textbf{19.33} & 25.06 \\
UniLM+\textsc{MediaSum} & \textbf{51.90} & \textbf{19.33} & \textbf{25.58} \\ \thickhline
\rowcolor[gray]{0.95}
\multicolumn{4}{c}{ICSI}                                       \\ \thickhline
UniLM             & 42.91 & 9.78  & 17.72 \\
UniLM+\textsc{MediaSum} & \textbf{43.65} & \textbf{10.13} & \textbf{18.59} \\ \thickhline
\rowcolor[gray]{0.95}
\multicolumn{4}{c}{SAMSum}                                     \\ \thickhline
UniLM             & 50.00 & 26.03 & 42.34 \\
UniLM+\textsc{MediaSum} & \textbf{50.55} & \textbf{26.39} & \textbf{42.68} \\ \thickhline
\end{tabular}
    \caption{Results on AMI, ICSI and SAMSum by using \textsc{MediaSum} as a dataset for transfer learning.\label{tab:transfer}}
\end{table}

\begin{table*}[ht]
\begin{tabular}{l}
\thickhline
\{\\
\hspace{0.3cm}"id": "NPR-11",\\
 \hspace{0.3cm}"program": "Day to Day",\\
  \hspace{0.3cm}"date": "2008-06-10",\\
  \hspace{0.3cm}"url": "https://www.npr.org/templates/story/story.php?storyId=91356794",\\
  \hspace{0.3cm}"title": "Researchers Find Discriminating Plants",\\
  \hspace{0.3cm}\makecell{"summary": "The 'sea rocket' shows preferential treatment to plants that are its kin. Evolutionary plant\\ ecologist Susan Dudley of McMaster University in Ontario discusses her discovery.",}\\
  \hspace{0.3cm}"utt": [\\
    \hspace{0.6cm}"This is Day to Day.  I'm Madeleine Brand.",\\
    \hspace{0.6cm}"And I'm Alex Cohen.",\\
    \hspace{0.6cm}"Coming up, the question of who wrote a famous religious poem turns into a very unchristian battle.",\\
    \hspace{0.6cm}\makecell{"First, remember the 1970s?  People talked to their houseplants, played them classical music. They \\were convinced plants were sensuous beings and there was that 1979 movie, 'The Secret Life of Plants.'",}\\
    \hspace{0.6cm}\makecell{"Only a few daring individuals, from the scientific establishment, have come forward with offers to \\replicate his experiments, or test his results. The great majority are content simply to condemn his efforts \\ without taking the trouble to investigate their validity.",}\\
    \hspace{0.6cm}...\\
    \hspace{0.6cm}"OK. Thank you.",\\
    \hspace{0.6cm}\makecell{"That's Susan Dudley. She's an associate professor of biology at McMaster University in Hamilt on \\Ontario. She discovered that there is a social life of plants."}\\
  \hspace{0.3cm}],\\
  \hspace{0.3cm}"speaker": [\\ 
    \hspace{0.6cm}"MADELEINE BRAND, host",\\
    \hspace{0.6cm}"ALEX COHEN, host",\\
    \hspace{0.6cm}"ALEX COHEN, host",\\
    \hspace{0.6cm}"MADELEINE BRAND, host",\\
    \hspace{0.6cm}"Unidentified Male",    \\
    \hspace{0.6cm}...\\
    \hspace{0.6cm}"Professor SUSAN DUDLEY (Biology, McMaster University)",\\
    \hspace{0.6cm}"MADELEINE BRAND, host"\\
  \hspace{0.3cm}]\\
 \} \\
\thickhline
\end{tabular}
    \caption{Example dialogue and summary from \textsc{MediaSum}. The number of strings in \textit{utt} and \textit{speaker} fields are the same. \label{tab:example}}
\end{table*}

\section{Conclusion}
We introduce \textsc{MediaSum}, a large-scale media interview dataset for dialogue summarization, consisting of 463.6K transcripts and summaries from NPR and CNN. We conduct transcript segmentation to align topic descriptions to segments for CNN interviews. The \textsc{MediaSum} dataset is an order of magnitude larger than existing corpora and contains complex multi-party conversations from multiple domains. We also show that \textsc{MediaSum} can be used as a dataset for transfer learning to improve a model's performance on other dialogue summarization tasks.

\section*{Ethics}
We have used only the publicly available transcripts data from the media sources and adhere to their only-for-research-purpose guideline.

As media and guests may have biased views, the transcripts and summaries will likely contain them. The content of the transcripts and summaries only reflect the views of the media and guests, and should be viewed with discretion. 

\section*{Acknowledgement}
We thank William Hinthorn for proof-reading the paper and thank the anonymous reviewers for their insightful comments.

\bibliography{main}
\bibliographystyle{acl_natbib}

\clearpage
\appendix

\begin{figure*}[!h]
    \centering
    %\vspace{-0.5cm}
    \includegraphics[trim=0cm 0cm 0cm 0cm, clip, width=0.95\linewidth]{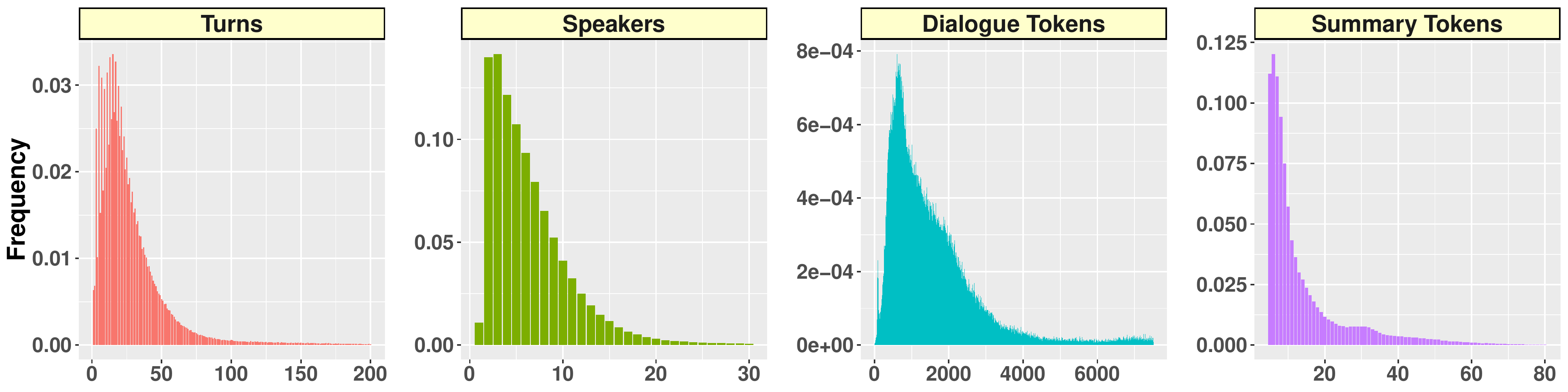}
    %\vspace{-4cm}
    \caption{Distribution of the number of turns, speakers, dialogue words and summary words in the dialogues of \textsc{MediaSum} dataset.}
    \label{fig:appendix_stat}
\end{figure*}

\begin{figure}[h]
    \centering
    %\vspace{-0.5cm}
    \includegraphics[trim=0cm 0cm 0cm 0cm, clip, width=0.95\linewidth]{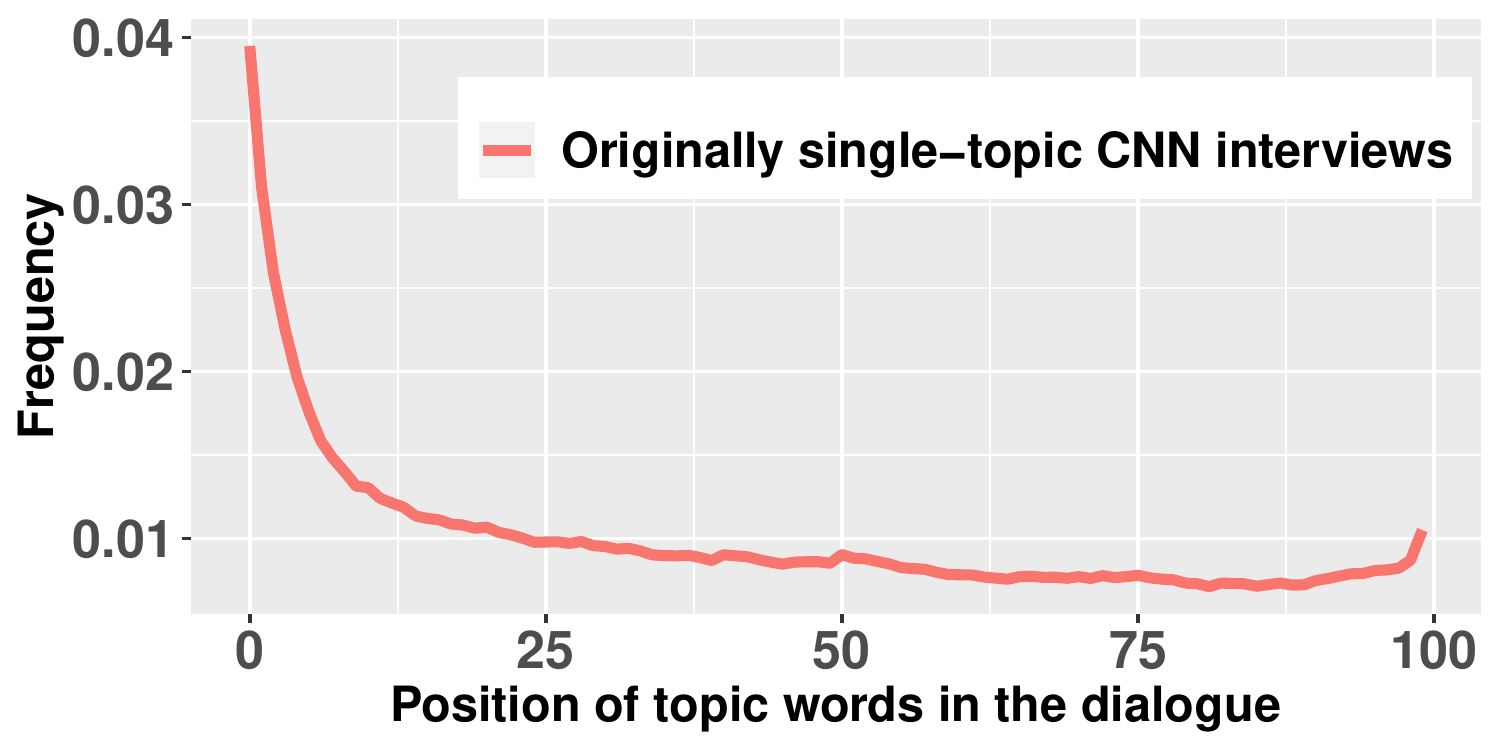}
    %\vspace{-4cm}
    \caption{The frequency of non-stop topic words appearing in different positions of the dialogue. The dialogues are from the original CNN transcripts with one topic. The positions are normalized to [0, 100].}
    \label{fig:poscnn_sg}
\end{figure}

% \begin{figure*}[h!]
%     \centering
%     %\vspace{-0.5cm}
%     \includegraphics[trim=0cm 0cm 0cm 0cm, clip, width=0.7\linewidth]{npr_example.png}
%     %\vspace{-4cm}
%     \caption{Example NPR transcript with summary (from \url{https://www.npr.org/2020/11/07/932422798/covid-19-cases-in-u-s-reach-record-highs-again}).}
%     \label{fig:npr_example}
% \end{figure*}

% \begin{figure*}[h!]
%     \centering
%     %\vspace{-0.5cm}
%     \includegraphics[trim=0cm 0cm 0cm 0cm, clip, width=0.7\linewidth]{cnn_example.png}
%     %\vspace{-4cm}
%     \caption{Example CNN transcript with topic descriptions separated by semicolons (from \url{http://transcripts.cnn.com/TRANSCRIPTS/2011/02/cnr.06.html}).}
%     \label{fig:cnn_example}
% \end{figure*}

%\section{Example}
%Fig.~\ref{fig:npr_example} and Fig.~\ref{fig:cnn_example} show example transcript pages from NPR and CNN. We use the overview text in NPR and topic descriptions in CNN as summaries.

\section{Data statistics}
\label{sec:appendix_stat}
Fig.~\ref{fig:appendix_stat} shows the distribution of the number of turns, speakers, dialogue words and summary words in the dialogues of \textsc{MediaSum} dataset. As shown, most dialogues have more than 500 words and 2 to 5 speakers.

\section{Topic analysis}
\label{sec:appendix_topic}
Table~\ref{tab:lda} shows the top 10 words in each cluster of \textsc{MediaSum} dialogues computed by the Latent Dirichlet Allocation tool in scikit-learn package.

\begin{table*}[t]
    \centering
    \begin{tabular}{c|c}
    \thickhline
        \textbf{Cluster} & \textbf{Top 10 words} \\
        \hline
        1 & prime, gop, iraq, bush, president, secretary, clinton, south, minister, white \\
        2 & plane, gop, today, look, report, rep, libya, crash, flight, continues \\
        3 & obama, coronavirus, attack, school, big, toll, saudi, gas, war, prices \\
        4 & forces, war, qaeda, crisis, syria, attack, middle, new, east, iraq \\
        5 & jobs, campaign, russian, news, white, tax, interview, old, president, iran \\
        6 & virginia, dead, new, suspect, day, case, covid, murder, 19, death \\
        7 & election, police, supreme, democrats, vote, house, impeachment, new, china, care \\
        8 & report, york, cnn, sanders, candidates, race, biden, democratic, president, presidential \\        
        \thickhline
    \end{tabular}
    \caption{Top 10 topics words in each cluster of \textsc{MediaSum} dialogues computed by the Latent Dirichlet Allocation tool in scikit-learn package. \label{tab:lda}}
\end{table*}

\section{Positional bias}
\label{sec:appendix_posbias}
Fig. \ref{fig:poscnn_sg} shows the frequency of non-stop topic words appearing in different positions of the dialogue. The dialogues are from the original CNN transcripts with one topic. The trend is mostly similar to that in Fig.~\ref{fig:pos}, except for a slight increase near the end. Thus, it shows that in televised programs, most topic keywords are mentioned at the beginning.

\section{Implementation Details}
\label{sec:appendix_imp}
For BART \citep{bart}, we use a learning rate of $2\times10^{-5}$, a batch size of 24 and train for 10 epochs. During beam search, we use a beam width of 3, and limits the minimum/maximum length of generated summary to be 3 and 80 tokens, respectively. The result on validation set of \textsc{MediaSum} is: 35.01 in ROUGE-1, 17.92 in ROUGE-2 and 31.15 in ROUGE-L.

For PTGen \citep{pgnet}, we use a vocabulary of 50,000 words. The model is a LSTM-based encoder-decoder model with a hidden size of 512. We train the model with Adagrad optimizer for 10 epochs and a learning rate of 0.1.
The result on validation set of \textsc{MediaSum} is: 28.07 in ROUGE-1,	12.11 in ROUGE-2 and 23.40 in ROUGE-L.

For UniLM \citep{UniLM}, we train the model with Adam optimizer for 100,000 steps with 2,000 warmup steps and learning rate is set to $1.5\times10^{-5}$. 
The result on validation set of \textsc{MediaSum} is: 32.27 in ROUGE-1, 16.99 in ROUGE-2 and 29.06 in ROUGE-L.

In all experiments, we truncate the input after 1,024 tokens. We use 8 v100 GPUs for the computation.

We follow \citet{hmnet} to adopt 100/17/20 and 43/10/6 for train/dev/test split on AMI and ICSI respectively. We employ the split for SAMSum following \citet{samsum}.

\section{Results on partitions}
\begin{table}[hbt!]
    \centering
    \begin{tabular}{l|c|c|c}
    \thickhline
        \textbf{Model} & \textbf{R-1} & \textbf{R-2} & \textbf{R-L}
        \\
    %     \thickhline
    %     \rowcolor[gray]{0.95} \multicolumn{4}{c}{Combined}\\
    %     \thickhline
    %     %LEAD-3 & 14.96 & 5.10 & 13.29 \\
    %     LEAD-3 &14.96&5.10&12.03\\
    %     %LEAD-2-1 & 13.25 & 4.35 & 11.80 \\
    %     \hline
    %     PTGen &28.77&12.24&24.18 \\
    %   % BART & 35.13 & 18.00 & 31.97\\
    %   BART & \textbf{35.09}&\textbf{18.05}&\textbf{31.44}\\
    %     UniLM &32.70&17.27&29.82\\

        \thickhline
        \rowcolor[gray]{0.95} \multicolumn{4}{c}{CNN}\\
        \thickhline
        %LEAD-3 & 13.35 & 4.37 & 11.94 \\
        LEAD-3 &13.36&4.37&11.10\\
        %LEAD-2-1 & 13.25 & 4.35 & 11.80 \\
        \hline
        PTGen &27.54&11.47&23.45 \\
        %BART & 34.08 & 17.53 & 31.41\\
                        BART & 34.07&17.57&31.36\\

        %BART$_{Com}$ & 34.10 & 17.50 & 31.23 \\
        UniLM & 31.97&16.97&29.88\\
        UniLM$_{Com}$ & 31.88&16.97&29.79\\

        \thickhline
        \rowcolor[gray]{0.95} \multicolumn{4}{c}{NPR}\\
        \thickhline
        %LEAD-3 & 28.43 & 11.23 & 24.58 \\
        LEAD-3 & 28.39&11.21&19.90\\
        \hline
        PTGen &35.86&16.01&24.46 \\
        %BART & 44.21 & 22.60 & 38.38\\
        BART & 43.55&21.99&32.03\\

        %BART$_{Com}$ & 43.72 & 22.11 & 38.05 \\
        UniLM & 41.42&20.73&30.65\\
        UniLM$_{Com}$ & 41.58&21.25&31.24\\

        \thickhline
    \end{tabular}
    \caption{ROUGE-1, ROUGE-2 and ROUGE-L F1 scores on the CNN and NPR partitions of the test data.  All models are trained on the corresponding partition of the training data, except UniLM$_{Com}$, which is trained on the entire \textsc{MediaSum}.  \label{tab:appendix_res}}
\end{table}

Table~\ref{tab:appendix_res} shows the results of models on the CNN and NPR partitions of the test data. All models are trained on the corresponding partition of the training data, except UniLM$_{Com}$, which is trained on the entire \textsc{MediaSum}. 

First, we notice that the result on NPR partition are better than that on CNN partition. Secondly, training on \textsc{MediaSum} can improve the ROUGE-L score by 0.6\% on NPR partition, compared with using NPR partition only for training.

\end{document}